\documentclass{article}

\usepackage{PRIMEarxiv}

\usepackage[utf8]{inputenc} 
\usepackage[T1]{fontenc}    
\usepackage{hyperref}       
\usepackage{url}            
\usepackage{booktabs}       
\usepackage{amsfonts}       
\usepackage{nicefrac}       
\usepackage{microtype}      
\usepackage{lipsum}
\usepackage{fancyhdr}       
\usepackage{graphicx}       
\graphicspath{{media/}}     
\usepackage{subcaption} 
\usepackage{algorithm}
\usepackage{algorithmic}
\usepackage{changepage}
\usepackage{pdfcomment}
\usepackage{xcolor}
\newcommand{\rev}{\textcolor{gray}}

\title{CL-Polyp: Contrastive Learning for Accurate Polyp Segmentation
}

\author{
  Desheng \rev{Li}, Chaoliang \rev{Liu} \\\\
  School of Artificial Intelligence and Computer Science 
  Jiangnan University \\
  Wuxi, 214122, China\\
   \And
  Zhiyong \rev{Xiao*} \\\\
  School of Artificial Intelligence and Computer Science 
  Jiangnan University \\
  Wuxi, 214122, China \\
  \texttt{zhiyong.xiao@jiangnan.edu.cn} \\
}

\begin{document}
\maketitle

\begin{abstract}
\label{}
Accurate segmentation of polyps from colonoscopy images is crucial for the early diagnosis and treatment of colorectal cancer. Most existing deep learning-based polyp segmentation methods adopt an Encoder-Decoder architecture, and some utilize multi-task frameworks that incorporate auxiliary tasks like classification to improve segmentation. However, these methods often need more labeled data and depend on task similarity, potentially limiting generalizability.
To address these challenges, we propose CL-Polyp, a contrastive learning-enhanced polyp segmentation network. Our method uses contrastive learning to enhance the encoder’s extraction of discriminative features by contrasting positive and negative sample pairs from polyp images. This self-supervised strategy improves visual representation without needing additional annotations. We also introduce two efficient, lightweight modules: the Modified Atrous Spatial Pyramid Pooling (MASPP) module for improved multi-scale feature fusion, and the Channel Concatenate and Element Add (CA) module to merge low-level and upsampled features for {enhanced} boundary reconstruction.
Extensive experiments on five benchmark datasets—Kvasir-SEG, CVC-ClinicDB, CVC-ColonDB, CVC-300, and ETIS—show that CL-Polyp consistently surpasses state-of-the-art methods. Specifically, it enhances the IoU metric by 0.011 and 0.020 on the Kvasir-SEG and CVC-ClinicDB datasets, respectively, demonstrating its effectiveness in clinical polyp segmentation.
\end{abstract}


\section{Introduction}
Colorectal cancer (CRC) is the third most common cancer globally, with a high incidence of colorectal adenomatous polyps being the primary manifestation\cite{Siegel2020}.
In clinical practice, colonoscopy is the gold standard for identifying diseased tissues in the gastrointestinal tract. Accurate polyp segmentation in colonoscopy images provides essential information for colorectal cancer treatment.

With advances in computer vision, image semantic segmentation is now widely used for polyp segmentation, simplifying clinicians' work. It also enhances medical image analysis and provides more valuable information for clinical practice and pathological research.

Polyp segmentation is also a challenging medical image segmentation task, mainly due to the characteristics of colonoscopy, gastrointestinal tract, and polyps\cite{bhalerao2023clustering}. It mainly includes:
1. Irregular size and shape of polyps;
2. Polyps often have similar color to the surrounding intestine, making it difficult to distinguish them from mucosa or residual digestive matter, complicating edge separation.
3. Polyp segmentation datasets are often small due to the high cost of manual annotation.
Existing deep learning methods are gradually solving these problems.

\begin{figure}[!ht]
	\centering	
	\includegraphics[width=0.7\linewidth]{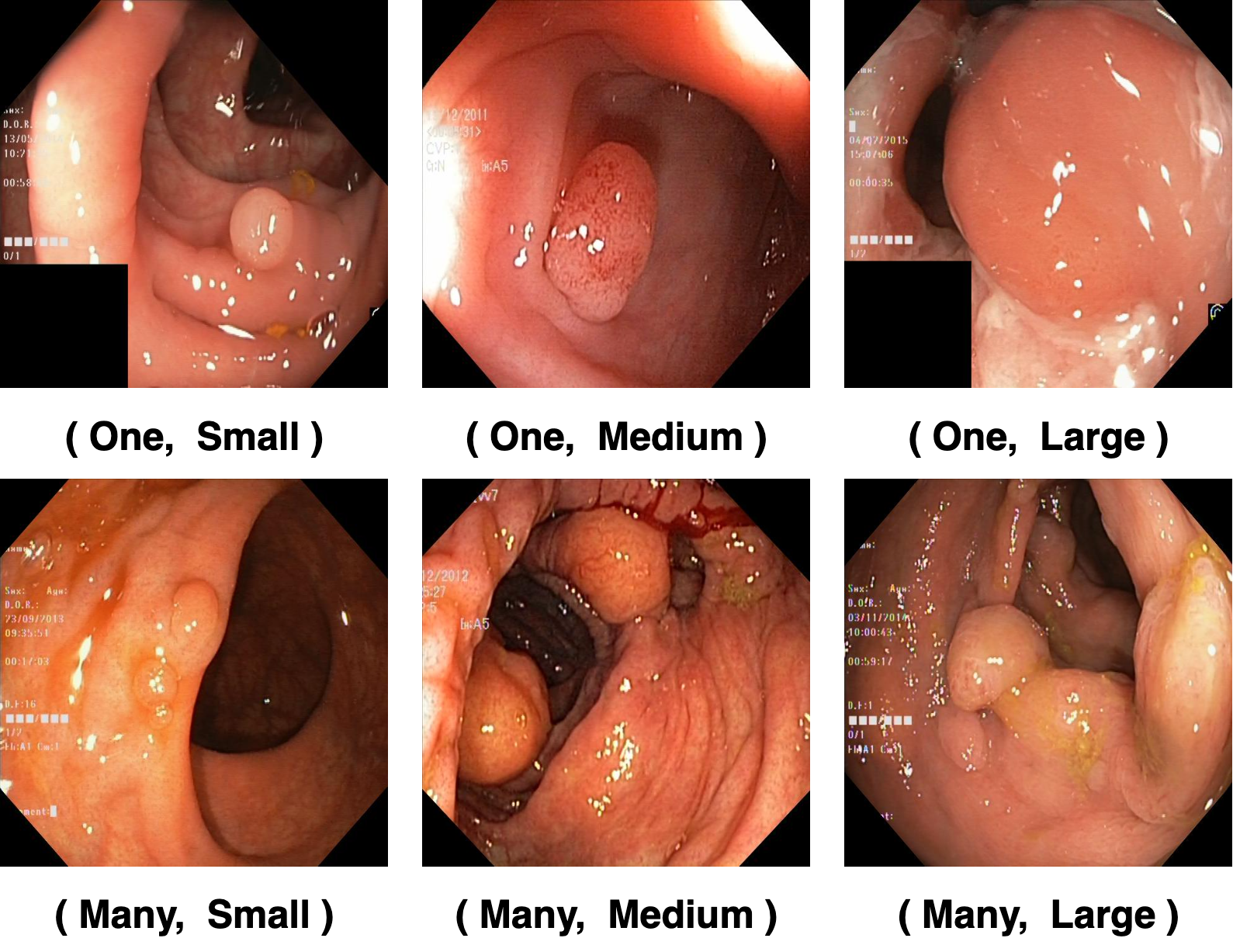}
	\caption{Polyp images from the Kvasir-SEG dataset\cite{DBLP:conf/mmm/JhaSRHLJJ20}, shown according to the classification method of TGA-Net\cite{DBLP:conf/miccai/TomarJBA22}.}
	\label{tga}
\end{figure}

Recent polyp segmentation methods using CNN and/or Transformer, and mainly adopting the Encoder-Decoder architecture, have achieved significant success.
The Encoder-Decoder architecture has been widely used in medical image segmentation since U-Net, and almost all polyp segmentation methods use this structure.
For clarity, we categorize these Encoder-Decoder architectures into three types based on the emphasis of the model structure (or the relative complexity of the Encoder and Decoder).

Architecture with equal emphasis on Encoder and Decoder, such as U-Net\cite{DBLP:conf/miccai/RonnebergerFB15}, U-Net++\cite{DBLP:journals/tmi/ZhouSTL20}, and ResU-Net ++\cite{DBLP:conf/ism/JhaSRJLHJ19} with a U-shaped structure, usually under the Encoder and Decoder, uses a similar design. It transmits the features of different Encoder layers to Decoder through skip connection, generally using element addition or channel concatenation. 
The information obtained from the various layers of the encoder is as crucial as the information recovered from the decoder. This paper presents a constrained nonnegative matrix factorization (NMF)-based method for hyperspectral image dimensionality reduction \cite{RSL2014, Xiao2014}. Hybrid Parallel Genetic Algorithm addresses the limitations of traditional segmentation methods—such as dependence on initial values, susceptibility to local optima, and slow computation—through parallel evolution and global optimization, making it particularly suitable for automated segmentation tasks involving high-dimensional, high-resolution, and multi-modal medical images \cite{HPGA2021}.

But more models focus on the structure of the decoder, usually using pre-trained BackBone as the encoder to extract features, but typically pay more attention to skip connections and decoder to improve the boundary information of polyps.
It is used to recover the boundary information of polyps, such as Psi-net\cite{DBLP:conf/embc/MurugesanSSRJS19}, PraNet\cite{DBLP:conf/miccai/FanJZCFSS20}, SA-Net\cite{DBLP:conf/miccai/WeiHZLZC21}, LDNet\cite{DBLP:conf/miccai/ZhangLWFGWL22}, and other structures.

Focusing on the structure of the encoder, the design of CNN and/or Transformer is usually used to extract features, and the simple form of the decoder is used to recover detailed information. Such as DeepLabV3+ \cite{DBLP:conf/eccv/ChenZPSA18}, TransFuse \cite{DBLP:conf/miccai/ZhangLH21}, and Polyp-Pvt \cite{DBLP:journals/corr/abs-2108-06932}. I think the focus of the model is the extraction of Encoder features. DeepLabV3+ uses ResNet \cite{DBLP:conf/cvpr/HeZRS16} to extract features. Polyp-Pvt uses PvT\cite{DBLP:conf/iccv/WangX0FSLL0021} to extract features.
CNN-ViT hybrid architectures have been widely adopted for image analysis\cite{Xiao2024FoodViT,Xiao2024FoodSwinViT,XIAO2025FoodCSWin,XIAO2025FGFoodNet,XIAO2025pestclassification,Wang2025}.

Recently, the structure of the multi-task model has come into our vision\cite{bhalerao2023essdm}. In the Encoder-Decoder structure, more branches are used to do other tasks, and the most commonly used are classification tasks, such as FCP-Net \cite{DBLP:journals/tmi/LiuZLZHFD22} and TGANet\cite{DBLP:conf/miccai/TomarJBA22}.
In addition to the segmentation task and classification task, FCPNet also has the interaction task of classification and segmentation. TGA-Net classifies polyps according to number and size to optimize Decoder(as shown in Fig.\ref{tga}).
The primary role of the multi-task model is to optimize the Encoder through multiple tasks to obtain better features.  However, the problem with this is that other functions in the multi-task may not be more effective for the Encoder.  The classification task is too single and too specific, and simple classification will not produce good results for segmentation \cite{kumar2014detection}.  
since multi-task models usually require additional label information, such as classifying benign and malignant tumors, it requires more manually annotated information and increases the workload of medical staff \cite{bhalerao2024imagined}.

In order to improve the accuracy and robustness of medical image segmentation, a medical image segmentation method based on improved convolutional neural network is proposed \cite{Xiao2019, Qian2020,Liu20223}. A deep learning-based method called SR-Net combines a bidirectional recurrent convolutional network (S-Net) for inter-slice correlation and a refinement network (R-Net) for intra-slice detail enhancement, along with deformable convolutions and data consistency operations, to improve the quality and speed of undersampled multislice MR image reconstruction \cite{CMPB2021}.
This paper proposes a lightweight multi-view hierarchical split network (MVHS-Net) for brain tumor segmentation in MRI images, which achieves high accuracy with significantly reduced computational complexity compared to existing methods \cite{BSPC2021}. This paper proposes a lightweight CNN-MLP hybrid model called RMMLP, which uses rolling tensors and matrix decomposition to effectively capture local and global features for accurate skin lesion segmentation, achieving state-of-the-art results with fewer parameters compared to Transformer-based methods \cite{BSPC2023}.
This paper proposes a novel semi-supervised medical image segmentation method that integrates a dual-teacher architecture—combining CNN and Transformer models—with uncertainty-guided training to effectively leverage both labeled and unlabeled MRI data, achieving superior segmentation performance using limited annotations \cite{XIAO2022107099}.
This paper proposes a lightweight 3D hippocampus segmentation method called Light3DHS, which combines a multi-scale convolutional attention module and a 3D MobileViT block to effectively fuse local and global features, achieving superior segmentation performance with fewer parameters and lower computational cost on multiple public datasets \cite{XIAO2024120608}.

As the demand for early diagnosis and treatment of colorectal cancer increases, the accuracy of polyp segmentation becomes particularly important. Currently, traditional deep learning methods have achieved some success in segmentation tasks, but they still face challenges such as small dataset sizes, high annotation costs, and the diverse shapes of polyps. These factors make it difficult for models to achieve ideal performance in practical applications. Therefore, exploring new learning methods to enhance segmentation accuracy, especially through contrastive learning to improve the model's sensitivity to polyp features, has become a key motivation for this research. To address the above challenges, we propose adding a contrastive learning-related visual representation learning task in training the polyp segmentation model, called CL-Polyp. Contrastive learning is a self-supervised learning approach that emphasizes learning representations by contrasting positive and negative pairs. In the context of polyp segmentation, contrastive learning can enhance the model's ability to discern subtle differences between polyp and non-polyp regions. By maximizing the similarity between positive samples—such as different views or augmentations of the same polyp—and minimizing the similarity between negative samples, the model can develop a more robust feature representation. This capability is particularly beneficial in medical imaging, where variability in polyp appearance can complicate segmentation tasks.  Specifically, we add a contrastive learning task to the polyp segmentation network. According to the classification standards of the number and size of polyps in TGANet,
Positive and negative samples for the contrastive learning task are selected and subjected to reasonable data augmentation. The polyp image is passed through the encoder to extract semantic information, and the encoder with momentum update is used to extract the information of positive and negative sample pairs. The entire Encoder-Decoder model is optimized by minimizing both the contrastive learning and segmentation loss functions.

CL-Polyp is similar to the multi-task segmentation model but different from the specific multi-task model, the model can implicitly learn features to optimize the segmentation results through contrastive learning, and better visual representation can be obtained. In addition, to better use the features in Encoder, we also modify the SOTA model DeepLabV3+. We propose the MASPP module, which can better fuse the deep multi-scale features. For recovering polyp boundary information, we presented the channel Concate and element Add (CA module) to make a better fusion between the low-level feature and upsampling feature, which can reconstruct the polyps' boundary information. Finally, we validate our model on five challenging polyp segmentation datasets: Kvasir-SEG, CVC-ClinicDB, CVCColonDB, CVC-300, and ETIS. Specifically, on the Kvasir-SEG dataset, our method achieves a mean Dice of 0.918 and an IoU of 0.864. On the CVC-ClinicDB dataset, our approach achieves a mean Dice of 0.955 and an IoU of 0.915.

To summarize, the main contributions of this paper are as follows:
\begin{itemize}
\item We propose a novel polyp segmentation framework that incorporates contrastive learning to optimize the encoder. Unlike traditional multi-task models, our approach improves feature representation without requiring additional annotation, enhancing the encoder’s learning capability in a self-supervised manner.
\item We design two lightweight yet effective modules to improve segmentation performance: the Modified Atrous Spatial Pyramid Pooling (MASPP) module for enhanced multi-scale feature fusion, and the Channel Concatenate and Element Add (CA) module for efficient integration of low-level and upsampled features.
\item Extensive experiments on five challenging polyp segmentation datasets demonstrate that our method surpasses existing state-of-the-art approaches in accuracy and robustness.
\end{itemize}

Next, in Section 2, we will briefly introduce the related work in the form of Fig.\ref{tga}. Then, we introduce our primary method in detail in Section 3, and the experimental session is in Section 4. Finally, Section 5 concludes and discusses future work.

\begin{figure*}[!ht]
    \centering
    \begin{subfigure}[b]{0.22\linewidth}
        \raisebox{0.21\height}{\includegraphics[width=\linewidth]{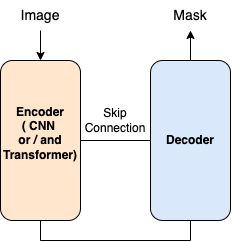}}
        \caption{Encoder and Decoder are equally important}
        \label{fig:a}
    \end{subfigure}
    \hfill
    \begin{subfigure}[b]{0.22\linewidth}
        \raisebox{0.21\height}{\includegraphics[width=\linewidth]{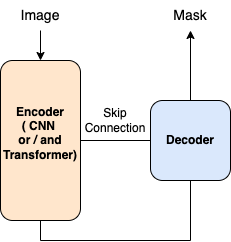}}
        \caption{Encoder is more critical than Decoder}
        \label{fig:b}
    \end{subfigure}
    \hfill
    \begin{subfigure}[b]{0.22\linewidth}
        \raisebox{0.21\height}{\includegraphics[width=\linewidth]{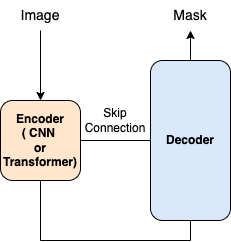}}
        \caption{The importance of the Decoder is higher than that of the Encoder}
        \label{fig:c}
    \end{subfigure}
    \hfill
    \begin{subfigure}[b]{0.22\linewidth}
        \includegraphics[width=\linewidth]{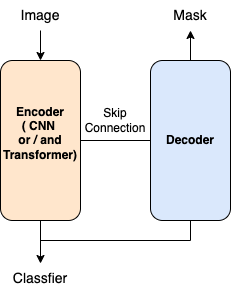}
        \caption{Multi-task model with an extra branch in the Encoder for classification tasks}
        \label{fig:d}
    \end{subfigure}
    \caption{Encoder-decoder structure variations for polyp segmentation models: (a) Balanced encoder and decoder; (b) Encoder with higher relative complexity; (c) Decoder with higher relative complexity; (d) Multi-task model with an auxiliary classification branch in the encoder.}
    \label{en-de}
\end{figure*}

\section{Related work}
\label{}
\subsection{Polyp segmentation}
Recently, numerous deep learning-based segmentation methods have been proposed, leading to broad applications and further development. 
FCN\cite{DBLP:conf/cvpr/LongSD15} is the first architecture to implement semantic segmentation end-to-end.
The proposed U-Net\cite{DBLP:conf/miccai/RonnebergerFB15} is a U-shaped symmetric method that uses an Encoder-Decoder structure and recovers image segmentation details through skip connections.
This simple and effective structure is widely used in medical image segmentation.
Based on this method, many researchers have optimized this benchmark.
We classify the structure of such encoder-decoders by the criteria of Encoder-Decoder relative size expressed as structural emphasis and/or relative complexity(As shown in Fig.\ref{en-de}).
The Encoder-Decoder architecture comprises an Encoder for generating feature maps and extracting high-level semantic information, a Decoder for recovering spatial information, and Skip Connections for fusing low-level and high-level features to enhance segmentation details (consider the skip connections as part of the Decoder).

1.The structure with equal emphasis on Encoder and Decoder is shown in Fig.\ref{en-de}a.  FCN, U-Net, U-Net++\cite{DBLP:journals/tmi/ZhouSTL20}, and ResU-net++\cite{DBLP:conf/ism/JhaSRJLHJ19} belong to this type of structure.
U-Net++ proposes dense skip connections and explores the effect of the multi-layer U-Net structure.
ResU-Net++ uses four techniques to optimize the segmentation network, which the are residual calculation, squeeze and excitation\cite{DBLP:conf/cvpr/HuSS18},  atrous spatial pyramidal pooling\cite{DBLP:conf/eccv/ChenZPSA18}, and attention mechanism\cite{DBLP:conf/nips/VaswaniSPUJGKP17}.

2.The structure of the model focused on Encoder is shown in Fig.\ref{en-de}b.
This type of structure focuses on extracting features through the Encoder, but this does not mean that the Decoder is not essential;  the Decoder is still a vital part.  The Encoder uses the architecture of CNN and/or Transformer.  The Transformer is supposed to be able to establish long-distance dependencies due to this property the Transformer.  Encoders using transformers tend to obtain more valuable and robust features, but this approach increases computation due to how MSA is calculated.  Some models use the Transformer as the Encoder to obtain enhanced features, fusing high- and low-level features through a fusion module.

DeeplabV3+\cite{DBLP:conf/eccv/ChenZPSA18}, Polyp-Pvt\cite{DBLP:journals/corr/abs-2108-06932}, HarDNet-MSEG\cite{DBLP:journals/corr/abs-2101-07172}, and TransFuse\cite{DBLP:conf/miccai/ZhangLH21} belong to this class of structures.
The DeepLabV3+ model uses ResNet\cite{DBLP:conf/cvpr/HeZRS16} as the Encoder model and uses the atrous convolution ASPP module to achieve the best effect on natural image semantic segmentation using only two upsampling.
The HarDNet-MSEG model employs the HarDBlk module as the Encoder to extract features.\
Polyp-PVT and TransFuse differ from DeepLabV3+ by using a Transformer structure as the Encoder.
Polyp-Pvt uses PvT\cite{DBLP:conf/iccv/WangX0FSLL0021} as Encoder and proposes a cascaded feature fusion module in Decoder, which achieves good results. 
TransFuse combines CNN and Transformer as the Encoder to extract spatial information and context dependencies.

3.The structure of the model focused on Decoder is shown in Fig.\ref{en-de}c. This kind of structure usually uses the existing Backbone structure to extract features. Still, the focus of the structure is on the Decoder part.
Compared with other models, the model focusing on the Decoder is the most. PraNet\cite{DBLP:conf/miccai/FanJZCFSS20}, SA-Net\cite{DBLP:conf/miccai/WeiHZLZC21}, MSNet\cite{DBLP:conf/miccai/ZhaoZL21}, CaraNet\cite{DBLP:journals/corr/abs-2108-07368} and GLFRNet\cite{DBLP:journals/tmi/SongCZSXCFPZ22} belong to this category.
PraNet uses Res2Net\cite{DBLP:journals/pami/GaoCZZYT21} as Encoder, adds an inverted attention module to the Decoder, and combines deep supervision to focus on the boundary area of polyps, which has a good effect.
SA-Net incorporates shadow attention in the Decoder to improve small polyp segmentation and proposes color exchange data augmentation to address overfitting.
BoxPolyp\cite{DBLP:conf/miccai/WeiHLCZL22} uses SA-Net as the baseline model and utilizes the bounding box utilization of a larger dataset, and solves a certain over-fitting problem by combining the bounding box with the model results.
MSNet adds an MS Module in the skip connection and Decoder, using multi-level and multi-stage cascaded subtraction operations to effectively obtain complementary information from lower to higher orders across different levels.
GLFRNet focuses on skip connections and the Decoder. The GFR module expands the receptive field and enriches low-level semantic information. The LFR module then performs upsampling and feature fusion.
A semi-supervised CT image segmentation method combines CNN and Transformer using entropy-constrained contrastive learning, which can improve segmentation accuracy by filtering unreliable pseudo-labels and leveraging both local and global feature information \cite{BEL2024}.

Some lightweight structures also belong to Fig.\ref{en-de}c. Such as FRCNet\cite{Shi2022}.
FRCnet introduces the ECC and PCF modules. Using a simple CNN structure as the Encoder, the ECC module is added to the Decoder to enhance fusion during up-sampling, while the PCF module is used for context fusion.

\subsection{Multi-task model}
Multi-task learning is a method to improve the learning ability of an encoder by learning tasks in parallel. In medical image segmentation, the relatively common segmentation task is combined with the classification task as shown in Fig.\ref{en-de}d. 
The primary purpose of this structure is to obtain better segmentation results through multi-task learning. CVNet\cite{DBLP:journals/mia/ZhouCLLXWYS21} uses the classification task to optimize the segmentation task in 3D chest ultrasound image segmentation to classify tumors into benign and malignant to optimize the learning ability of the Encoder.
FCPNet\cite{DBLP:journals/tmi/LiuZLZHFD22} proposes a task of feature compression pyramid network guided by game-theoretic interactions to conduct the segmentation task together with the classification task.
Recently, TGANet\cite{DBLP:conf/miccai/TomarJBA22} divided polyps into three categories according to number and size, one and multiple according to number, and small, medium and large according to size, and used the information obtained by classification as text attention guidance for segmentation.

Multi-task learning can enhance the Encoder without making many modifications to the model and without increasing the computational load of the model. However, the disadvantage of multi-task models is that other tasks have specific results, which may need to be more effective for segmentation. If not, it may even have the opposite effect. Furthermore, it requires additional label information, increasing the workload of medical staff.

\subsection{Contrastive learning}
Contrastive learning is the ability to learn generalization and transfer from unlabeled data. The general approach involves constructing positive and negative samples for each data point, aiming to reduce the distance to the positive sample.
At the same time, the distance between negative samples is increased. In this process, the selection of negative samples is the most important. MoCo\cite{DBLP:conf/cvpr/He0WXG20} uses dynamic memory to store negative samples in momentum encoder storage. 
SimCLR\cite{DBLP:conf/icml/ChenK0H20} shows that large batch sizes as well as long training times are also important for contrastive learning. By introducing a stopping gradient, SimSiam\cite{DBLP:conf/cvpr/ChenH21} indicates that meaningful representations can be learned from simple network architectures without negative pairs, large batch sizes, or momentum encoders.
Few researchers have used contrastive learning in polyp segmentation models, although this method is very effective.

\section{Methods}
In this paper, we propose that CL-Polyp consists of two branches: 
the segmentation and the contrastive learning branches.  
The main branch performs polyp segmentation, while contrastive learning serves as an auxiliary task. 
The implementation details are provided below.
\begin{figure*}[!ht]
	\centering	
	\includegraphics[width=0.9\linewidth]{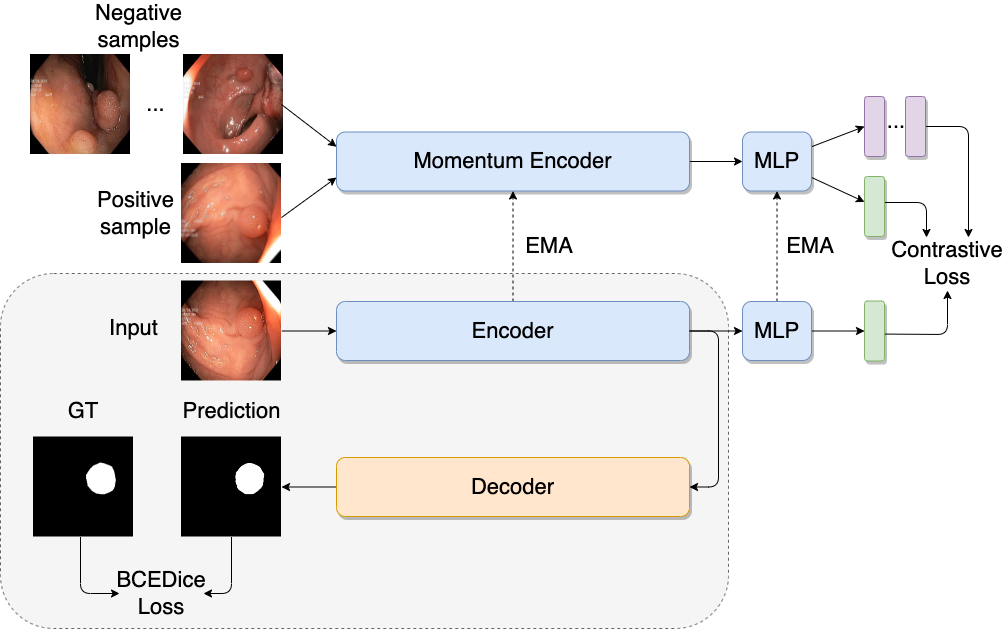}
	\caption{Architecture of the ResNet50-based segmentation model with contrastive learning: Feature maps extracted by the ResNet50 encoder are fed into a segmentation branch and a contrastive learning branch. The contrastive branch compares features with those from a momentum-updated encoder using positive and negative samples. The multi-layer perceptron (MLP) in the contrastive branch consists of linear, batch normalization (BN), and ReLU layers.}
	\label{method}
\end{figure*}
\subsection{Contrastive learning branch}
\subsubsection{Selection of positive and negative samples}
The training by contrastive learning is similar to the training method of the multi-task model structure, which uses a mixed loss function to train the network structure. Contrastive learning focuses on constructing positive and negative sample pairs for data. In this paper, according to the classification method of data in TGANet\cite{DBLP:conf/miccai/TomarJBA22}, polyps are divided into one or more according to the number of polyps and the size.
Based on polyp count, data can be divided into single or multiple polyps; based on size, they can be classified as small, medium, or large. 
Combining the number and size categories, the total can be divided into six categories, as shown in Fig.\ref{tga}.

$N$  and $S $   respectively, polyp number and size of the collection, polyp size, and the number of data sets described as polyps set $C = N \times S = \{(n, s) | n \in N, s \in S\}$. Training set data $D_ {tr} = \{(x, c) | c \in C \} $.
This categorization facilitates the selection of positive and negative samples.

Let's consider specific data for $D_ {tr} $ a data of $x^i $, for the description of the sample for $c = \{(n, s) | n = \hat {n}, s =  \hat s \} $. The number of polyps in the image can be represented as $\hat{n} $, and the size can be represented as $\hat{s} $.The description of the sample for $c = \{(n, s) | n = \hat {n}, s = \hat s \} $.

There are two ways to select positive samples $ x_p^i$ similar to the original image, one is data augmentation of image  and the other is from $c = \{(n, s) | n = \hat {n}, s = \hat s \} $ choose one of the data as the positive example. 
In our experiments, we use data augmentation of the image to create the positive sample.
The positive sample has the same number and size of polyps as the original data.
Negative selection and the image size and the number of samples of different samples, the samples of negative samples $ x_n^i $ to choose $\{(n, s) | n\ne   \hat {n}, s \ne  \hat {s} \} $.
\subsubsection{Detailed structure}
As shown in Fig.\ref{method},the Encoder comprises a ResNet and an MLP. Specifically, ResNet50 is used. The MLP consists of a Linear layer, BN layer, ReLU activation function, and Linear layer.
The image passes through ResNet's last layer block, then splits into two branches: one for contrastive learning and the other for the main segmentation branch.
After average pooling, the contrastive learning branch features are processed by the MLP to obtain a 2048 -dimensional vector. 
Specifically, the Encoder extracts visual features $h$ from the polyp image $x^i$,
Then, $x_p^i$ and $x_n^i$ are passed through a momentum encoder with the same structure to obtain $h_p$ and $h_n$, respectively.
The parameters of the momentum encoder are updated utilizing momentum in the following way.
\begin{equation}
\centering
\theta_k = m \theta_k + ( 1 - m ) \theta_q
\end{equation}
Then, we use {the} contrastive learning loss function to optimize these three feature vectors.
The overall training process is presented in Algorithm.\ref{alg1}.
For the inference procedure, as shown in the gray box in Fig.\ref{method}, there are no positive and negative samples and the process of momentum Encoder.

\begin{algorithm}
	\renewcommand{\algorithmicrequire}{\textbf{Input:}}
	\renewcommand{\algorithmicensure}{\textbf{Output:}}
	\caption{Pseudo code for the training procedure of CL-Polyp in a PyTorch-like style}
	\label{alg1}
	\begin{algorithmic}
		\REQUIRE input data $x^i$,  K negative samples $x_n^i$
		\ENSURE segmentation prediction 
		\STATE Initialize $F_q$, $F_k$: encoder networks for querry and key
		\STATE Initialize $D$ : decoder network
		\STATE $F_k$.params = $F_q$.params
		\FOR{$x$, GT, $K * x_n$ in loader}
		\STATE $x_q$, $x_p$ , $x_n$ = aug(x), aug(x), aug($x_n$)
		\STATE $h$,$h_p$, $h_n$ = $f_q$.forward($x_q$), $f_k$.forward($x_p$), $f_k$.forward($x_k$)
		\FOR{i in K}
		\STATE loss += $L_{CL}$($h$, $h_p$, $h_n$)
		\ENDFOR
		\STATE prediction = $D$.forward($h$)
		\STATE loss += $L_{SEG}$(GT, prediction)
		\STATE loss.backward()
		\STATE update($F_q$.params, $D$.params)
		\STATE $F_k.params = m*F_k.params + (1-m)*F_q.params $
		\ENDFOR 
	\end{algorithmic}  
\end{algorithm}

\subsubsection{Contrastive Learning Loss Function}
The contrastive learning method ensures that the features of a polyp image $X^i$ are similar to the positive sample $X_p^i$, such that the distance between image features and positive sample features is always less than the distance between negative sample features and image features.
The loss function of contrastive learning is introduced to optimize $h$, $h_p$, and $h_n$ to reduce the distance between $h$ and $h_p$.
Concurrently, the distance between $h$ and each negative sample $h_n$ is increased.
The triplet loss function is adopted, which is proposed in Face-Net\cite{DBLP:conf/cvpr/SchroffKP15} and the specific loss function can be described as
\begin{equation} L_{CL} = \sum_{i=1}^{N} \left[ \Vert f(x^i) - f(x_p^i) \Vert_{2}^{2} - \Vert f(x^i) - f(x_n^i) \Vert_{2}^{2} + \alpha \right] \end{equation}
Where $\alpha $ is the interval between positive and negative samples. $N $ is all instances of the training set.
In {practice}, {each} image corresponds to one positive sample and $K$ negative samples. In {our} {experiments}, $K$ is set to 4.
\subsubsection{Data Augmentation Techniques}
\begin{figure*}[!ht]
	\centering	
	\includegraphics[width=\linewidth]{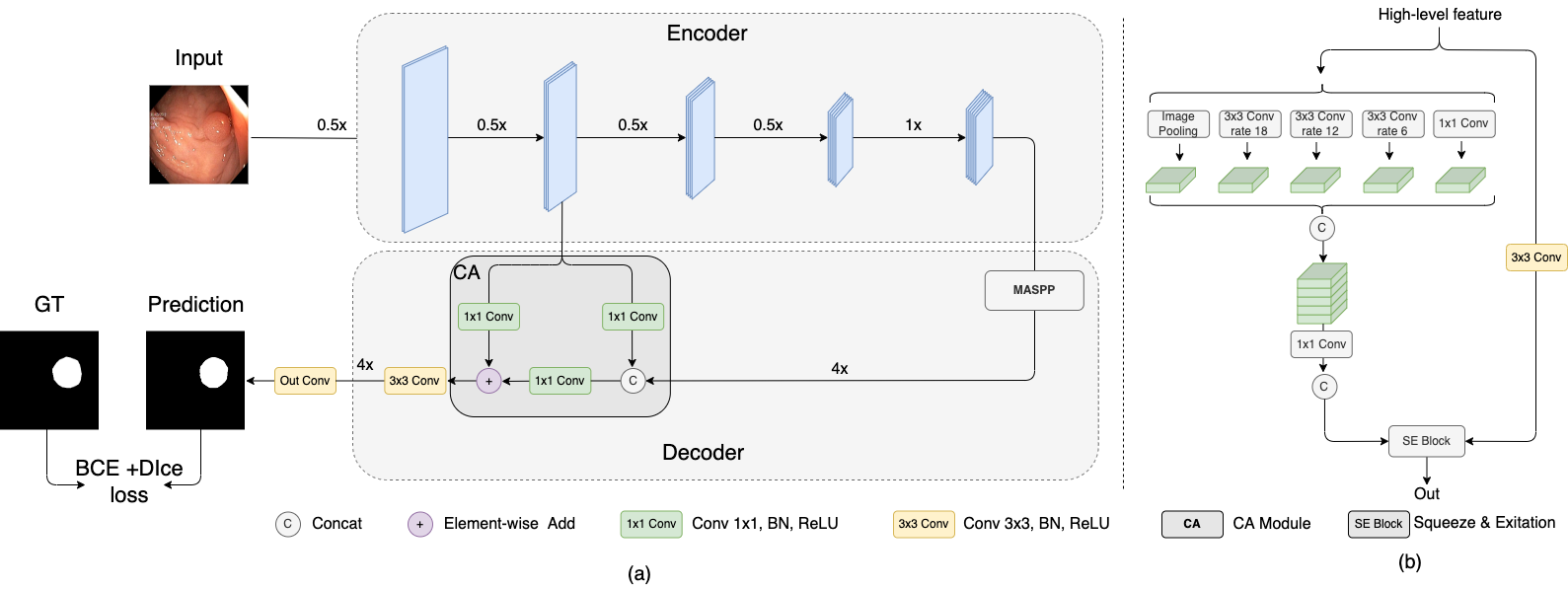}
	\caption{Segmentation model architecture: (a) Overall structure with ResNet50 encoder, MASPP, and CA modules in the decoder. (b) Detailed structure of the MASPP module.
}
	\label{model}
\end{figure*}

Learning data invariance is crucial for contrastive learning\cite{DBLP:journals/thms/BhaleraoP24}. Proper selection of data augmentation is essential for understanding data invariance, which directly impacts the model's learning ability. In polyp segmentation tasks, common data augmentation methods include image resizing and cropping, rotation, horizontal flipping, color transformation, and grayscale change. {We} selected random cropping, resizing, rotation, horizontal flipping, Gaussian noise, and normalization as our data augmentation techniques.

\subsection{Segmentation branch}
Unlike other models focusing on Decoder\cite{DBLP:journals/bspc/BhaleraoP22}, we adopt the DeepLabV3+ network structure as the Baseline model. Although the Baseline segmentation results are excellent, some things could still be improved so that we can achieve good results with only a few simple modifications to the Baseline model. We propose an enhanced ASPP module named MASPP module and use the CA module to improve the model. Our model belongs to the model that pays more attention to Encoder, as shown in Fig.\ref{en-de}b.Although the Decoder has been modified, the focus is still on the features obtained through the Encoder.
The detailed structure of our model can be seen in Fig.\ref{model}.
\subsubsection{Modified ASPP module}
The ASPP module proposed in DeppLabV3+, ASPP module can obtain multi-scale information through multiple dilated convolution modules with different proportions. In ASPP, channel concatenates, and 1x1 convolution layers are used to fuse multi-scale information after multiple dilated convolutions. Still, the fusion of multi-scale information can only partially be carried out by using channel concatenate. Therefore, we propose {the} Modified Atrous Spatial Pyramid Pooling (MASPP) module {to} {improve} {the} {fusion of} high-level multi-scale features.

Fig.\ref{model}b shows the detailed MASPP module. The high-level feature information $f_5$, and then two branch structures, the first branch structure is a ASPP module, after multiple dilated convolution modules, the channel concatenate and 1x1 convolution are simply fused. The other branch {maintains} the original feature structure {by passing} the features through 3x3 convolution layers, BN layers, and ReLU activation layers.
After that, an SE module, proposed in SENet\cite{DBLP:conf/cvpr/HuSS18}, is used to fuse these two branch features. The SE module includes Squeeze \& Excitation. In the Squeeze module,the first average pooling turn the feature $ f_5 \in \mathbb{R}^{C \times H \times W} $ into a shape of $ C \times 1 \times 1$. Then through a Linear layer, the ReLU activation layer, Linear layer becomes a shape of $ H \times W \times 1 \times 1 $ and then the original data and the activated vector are multiplied, which can better fuse multiple features. This can be expressed as follows.
\begin{equation}
	{f = \rm SE(\rm Conv(f_5), \rm ASPP(f_5))} 
\end{equation}
\subsubsection{CA Module}
In the DeepLabV3+ model, skip connections enhance boundary information using second -layer Decoder features. However, this ignores other layers. We propose the CA module to improve feature fusion in the Decoder.
The second layer feature $f_2 \in \mathbb{R}^{C \times H \times W} $ , $l$ represents the second number of layers of the feature, after two 1 x 1 convolution will output two features $A_i, B_i$, and the upsample feature $z_i $, concrete can be expressed as $z_i =  A_i + \rm Conv(\rm cat(B_i, z_i))$.
A 4-fold upsampling operation and an Outconv then obtain the final prediction result.
\subsubsection{Segmentation branch loss function}
To reduce overfitting, we adopt the BCEDice loss used in nnU-Net\cite{DBLP:journals/corr/abs-1809-10486}.
It consists of Dice loss and BCE loss.
Dice loss can be widely applied to medical image tasks and is written as follows.
\begin{equation}
	L_{Dice} =1 - 2\times  \frac{\sum _i P_i G_i}{ \sum _i P_i + \sum _i G_i} 
\end{equation}
BCE loss can be understood as follows.
\begin{equation}
	L_{BCE}  = - \sum_{i}[(1 - G_i)\ln(1 - P_{i}) + G_{i}\ln(P_{i})] 
\end{equation}
Where $i$ denotes the index of all pixels in the probability distribution map. $P_i$ is the probability that the $i$-th pixel belongs to the segmented region, and $G_i$ is the true value of the $i$-th pixel.
Finally, the {segmentation} loss function {combines} {BCE} {loss} and {Dice loss}.
\begin{equation}
	L_{Seg} =  L_{BCE}  + L_{Dice}
\end{equation}

\subsection{Hybrid loss function}
The overall loss function {combines} the contrastive learning {loss} and the {segmentation} loss.
\begin{equation}
L_{Total} = L_{Seg} +  \beta    L_{CL}
\end{equation}
Where $\beta$ is the hyper-parameter and is set to 0.5 in our experiments.

\section{Experiments and Results}
\subsection{Datasets and Scenarios}
To verify the effectiveness and accuracy of our method, we mainly use five datasets commonly used in polyp segmentation, which are respectively
Kvasir-SEG\cite{DBLP:conf/mmm/JhaSRHLJJ20}, CVC-CliniCDB\cite{DBLP:journals/cmig/BernalSFGRV15}, CVC-ColonDB\cite{DBLP:journals/tmi/TajbakhshGL16}, CVC-300\cite{DBLP:journals/corr/VazquezBSFLRDC16} and ETIS\cite{DBLP:journals/cars/SilvaHRDG14}.

The Kvasir-SEG dataset consists of 1000 images. The images in the dataset are irregular in size, ranging from 332x487 to 1920x1072.
The size of the polyp ranged from 0.79 \% to 62.13 \%, and the shape and size of the polyp were irregular.
So it's the most challenging.
The CVC-ClinicDB dataset includes 612 images obtained from 29 polyp videos. The image size is 384 x 288. 
The size varies from 0.34\% to 45.88\%.
The CVC-ColonDB dataset includes 380 images obtained from 16 colonoscopy videos with an image size of 574 x 500. Polyp size from
0.30\% to 63.15 \%.
The {CVC-300} dataset {is} a cross-domain dataset {containing} 60 polyp images.
The ETIS dataset includes 196 images obtained from 16 enteroscopy videos. 
The image size is 1225 x 996. It has a larger resolution compared to other images.

\textbf{Experiment Scenario I}: Our experiments on the Kvasir-SEG dataset alone. We split it into 700/300 for training and test sets instead of 800/100. The reason for the separate experiment is that the sample size of Kvasir-SEG is the most significant and more complex compared to other datasets.
The experimental results of our method on this data set can reflect the learning ability of the method.

\textbf{Experiment Scenario II}: We train according to the method of PraNet. PraNet is an extensive training exercise designed to test the generalization performance of a method.
Images from Kvasir-SEG and CVC-ClinicDB were randomly divided into 90\% for training and 10\% for testing. The training set contains 1450 images (900 Kvasir-SEG images and 550 CVC-ClinicDB images are 90\% randomly split from their respective datasets).
The remaining data from Kvasir-SEG, CVC-ClinicDB, CVC-ColonDB, CVC300, and ETIS datasets were used for testing, which was used to evaluate the generalization ability of the model.

\textbf{Experiment Scenario III}: 
We conduct ablation experiments on the Kvasi-SEG dataset. To verify the effectiveness of the various modules proposed in our method, the training set and test set are divided in the same way as that of Experiment Scenario I. Ablation experiments on the Kvaisr-SEG dataset can reveal the effectiveness of each part of our method. {Additionally}, we {experimented} to verify the impact of different data augmentations on the model.

\subsection{Experimental Setup}
The experiments use the PyTorch framework on Ubuntu22.04, using python version 3.9. The hardware device, using a single NVIDIA 1080TI GPU with 11GB memory, accelerates our segmentation model’s training with a mini-batch size of 4. Using the Adam optimizer, the model used an initial learning rate of 1e-4 to train our model. Learning Rate Decay Strategy Cosine learning rate decay strategy is chosen. Besides, the input images are resized to 384 x 384. Due to our training method, we trained {for} 300 epochs {in} {Experiment} {Scenarios} I, II, and III.

\subsection{Evaluation Metrics}
{Medical} image segmentation {commonly} uses the Dice coefficient and IoU as evaluation {metrics}. We used Dice and IoU as evaluation {metrics} in Experimental {Scenarios} I, II, and III. The {Dice} coefficient is a standard measure {for} {comparing} the predicted {segmentation} with the {ground} {truth} {and} can be defined as:
\begin{equation}
    Dice = \frac{2 \times TP}{2 \times TP + FP +FN} 
\end{equation}
IoU, another standard segmentation evaluation {metric}, {calculates} the overlap between the predicted and {ground truth} values, {as} {defined by} the following formula:
\begin{equation}
    IoU = \frac{TP}{TP + FP +FN} 
\end{equation}
TP, FP, TN, and FN represent True Positive, False Positive, True Negative, and False Negative respectively. 
{Additionally}, we compared three {metrics—}precision, recall, and {the} F2 score {—}in Experiment Scenario I.
\begin{equation}
    precision = \frac{TP}{TP + FP} 
\end{equation}
\begin{equation}
    recall = \frac{TP}{TP + FN}
\end{equation}
\begin{equation}
    F2 = \frac{5 \times Precision \times Recall}{4 \times Precision + Recall}
\end{equation}
\subsection{Experimental Results}
\subsubsection{Results of Experiment Scenario I}
In Experiment Scenario I, we compare our CL-Polyp with ten medical image segmentation methods, including U-Net\cite{DBLP:conf/miccai/RonnebergerFB15}, 
U-Net++\cite{DBLP:journals/tmi/ZhouSTL20},
DeepLabV3+\cite{DBLP:conf/eccv/ChenZPSA18},
PraNet\cite{DBLP:conf/miccai/FanJZCFSS20}, 
HarDNet-MSEG\cite{DBLP:journals/corr/abs-2101-07172}, 
TransFuse-S/L\cite{DBLP:conf/miccai/ZhangLH21}, 
TGANet\cite{DBLP:conf/miccai/TomarJBA22}, 
and  GLFRNet\cite{DBLP:journals/tmi/SongCZSXCFPZ22}. Among them, U-Net, U-Net++ and DeepLabV3+ are classic image segmentation methods; PraNet, HarDNet-MSEG, TransFuse-S/L, TGANet, and GLFRNet are recently advanced polyp  segmentation methods.

Table 1  shows the  results on the  Kvasir-SEG dataset.
The comparison of these models all achieved good performance, with Dice above 0.90 and IoU above 0.83. Some methods failed to reach the highest accuracy in the paper, because Experiment Scenario I was only conducted on the Kvasir-SEG dataset.
The PraNet method achieved a Dice of 0.894 and IoU of 0.830. The average Dice of TransFuse-S/L was 0.90 and IoU was 0.84.
GLFRNet (TMI'22) had the highest Dice of 0.911 and IoU of 0.853. The best result of our proposed method is that Dice reaches 0.918 and IoU reaches 0.864, which is higher than TGANet which uses the same partitioning method as us. {Our} {method} {also} {achieved} {the} {highest precision} and F2 {score}.

\begin{table}[]
\centering
\caption{{Segmentation} {performance} of ten methods on the Kvasir-SEG dataset {in Experiment Scenario I: Comparative results}.}
\resizebox{0.7\linewidth}{!}{
\begin{tabular}{lccccc}
\hline
Method                 & \multicolumn{5}{l}{Kvasir-SEG}                                                     \\ \cline{2-6} 
                       & Dice           & Iou            & Precision      & Recall         & F2             \\ \hline
U-Net(MICCAI'15)       & 0.826          & 0.747          & 0.870          & 0.850          & 0.835          \\
U-Net++(TMI'19)        & 0.830          & 0.753          & 0.863          & 0.834          & 0.833          \\
DeepLabV3+(ECCV'18)    & 0.889          & 0.827          & 0.912          & 0.906          & 0.898          \\
PraNet(MICCAI'20)      & 0.894          & 0.830          & 0.913          & 0.906          & 0.898          \\
HarDNet-MSEG(arXiv) & 0.904          & 0.848          & 0.907          & \textbf{0.923} & \textbf{0.915} \\
TransFuse-S(MICCAI'21) & 0.901          & 0.839          & 0.913          & 0.905          & 0.906          \\
TransFuse-L(MICCAI'21) & 0.902          & 0.844          & 0.914          & 0.905          & 0.906          \\
TGANet(MICCAI'22)      & 0.898          & 0.833          & 0.912          & 0.913          & 0.903          \\
GLFRNet(TMI'22)        & 0.911          & 0.853          & 0.927          & 0.900          & 0.913          \\
CL-Polyp(Ours)         & \textbf{0.918} & \textbf{0.864} & \textbf{0.946} & 0.910          & \textbf{0.915} \\ \hline
\end{tabular}}
\end{table}

\begin{figure*}[!ht]
	\centering	
	\includegraphics[width=0.8\linewidth]{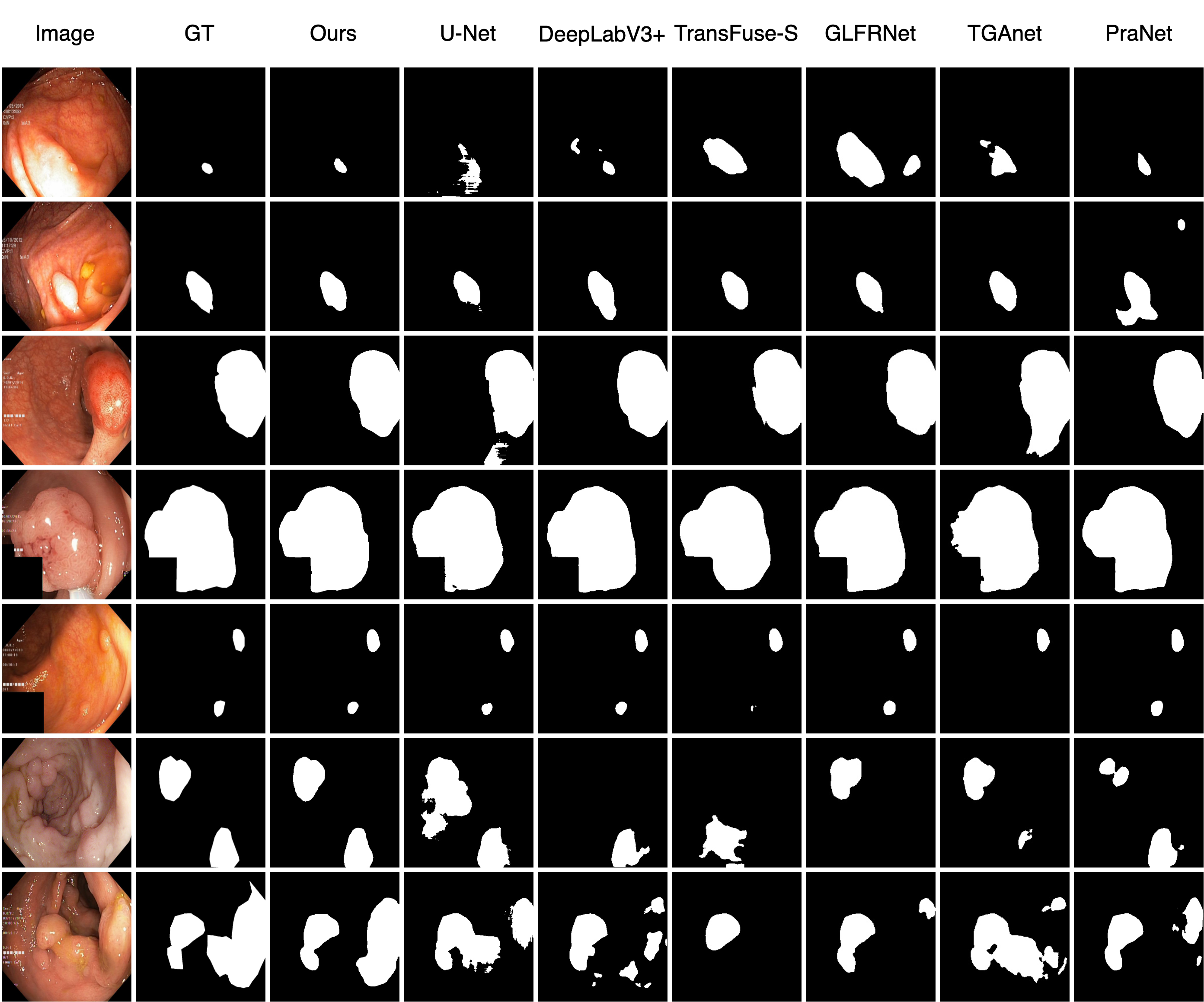}
	\caption{{Qualitative} comparison {of} {segmentation} {results: Proposed} method {versus} six state-of-the-art methods {on representative images}.}
	\label{result}
\end{figure*}

To show our model intuitively, we provide results of our model on Kvasir-SEG test images and compare testing results with six advanced models. As shown in Fig.\ref{result}, the proposed method can obtain better outcomes for the input image in the first four rows, regardless of the overall polyp or the edge details. Our approach can distinguish multiple polyps clearly for the figure's last three rows of pictures, while other models are challenging to determine. {Even} for polyps with fuzzy boundaries in the last row, our method effectively {segments} {them}.

\subsubsection{Results of Experiment Scenario II}
In Experiment Scenarios II, we compare our CL-Polyp with ten medical image segmentation methods, including U-Net, U-Net++,  PraNet, SA-Net\cite{DBLP:conf/miccai/WeiHZLZC21}, MSNet\cite{DBLP:conf/miccai/ZhaoZL21}, PolypPvt\cite{DBLP:journals/corr/abs-2108-06932}, BoxPolyp-Res2Net\cite{DBLP:conf/miccai/WeiHLCZL22} and LDNet\cite{DBLP:conf/miccai/ZhangLWFGWL22}. Among them, U-Net and  U-Net++ are classic image segmentation methods. PraNet, SA-Net, MSNet, PolypPvt, BoxPolyp-Res2Net, and LDNet are recently advanced polyp segmentation methods.

The results are shown in Table 2 and Table 3.
Results on Kvasir-SEG and CVC-ClinicDB. As we can see, 
our proposed model achieves a very high level on both data sets. Specifically, on the Kvasir-SEG dataset, Dice and Iou were 0.917 and 0.864, respectively. 
Our method performs almost as well as polyp-Pvt using Pvt as the Encoder on the Kvasir-SEG dataset.
We performed better in the CVC-ClinicDB dataset, with Dice and IoU reaching 0.955 and 0.915, respectively, exceeding the existing methods LDNet 1.2\% and 2\%, respectively.

Resluts on CVC-ColonDB, CVC-300 and ETIS.
 Our method only used these three data sets for testing, but not for training, to verify the method's generalization ability. It can be seen that although the performance of our approach on these three data sets is also excellent, it fails to reach the optimal index, and there is a gap of about 2 \% between our method and the other most advanced methods. This is due to our training method using contrastive learning, through which we can learn good visual representation. Still, to some extent, the model will be trapped in the current data set, which affects the model's generalization ability.

\begin{table}[]
\centering
\caption{{Segmentation performance} of {ten} methods on {the} Kvasir-SEG and CVC-ClinicDB datasets {using} the {PraNet} training method{: Comparative analysis} of {results}.}
\resizebox{0.7\linewidth}{!}{
\begin{tabular}{lcclcc}
\hline
Method                      & \multicolumn{2}{l}{Kvasir-SEG} &  & \multicolumn{2}{l}{CVC-ClinicDB} \\ \cline{2-3} \cline{5-6} 
                            & Dice           & IoU           &  & Dice            & IoU            \\ \hline
U-Net(MICCAI'15)            & 0.818          & 0.746         &  & 0.823           & 0.750          \\
U-Net++(TMI'19)             & 0.821          & 0.743         &  & 0.790           & 0.729          \\
PraNet(MICCAI'20)           & 0.898          & 0.840         &  & 0.899           & 0.849          \\
SA-Net(MICCAI'21)           & 0.904          & 0.847         &  & 0.916           & 0.859          \\
MSNet(MICCAI'21)           & 0.905          & 0.849          &  & 0.918           & 0.869          \\
Polyp-Pvt(arXiv)            & 0.917          & 0.864         &  & 0.937           & 0.889          \\
BoxPolyp-Res2Net(MICCAI'22) & 0.910          & 0.857         &  & 0.904           & 0.849          \\   
LDNet(MICCAI'22)            & 0.907          & 0.853         &  & 0.943           & 0.895          \\
CL-Polyp(Ours)              & \textbf{0.917}    & \textbf{0.864}      &  & \textbf{0.955 }  & \textbf{0.915} \\ \hline
\end{tabular}}
\end{table}

\begin{table}[]
\centering
\caption{{Segmentation performance} of ten methods on the CVC-ColonDB, CVC-300, and ETIS datasets {using} the {PraNet} training method{: Comparative analysis} of {results}. {Missing values are indicated by} ‘-’.}
\resizebox{0.7\linewidth}{!}{
\begin{tabular}{lcclccccc}
\hline
Method                                                                 & \multicolumn{2}{l}{CVC-ColonDB}                       &  & \multicolumn{2}{l}{CVC-300}                           &                      & \multicolumn{2}{l}{ETIS}                              \\ \cline{2-3} \cline{5-6} \cline{8-9} 
                                                                       & Dice                      & IoU                       &  & Dice                      & IoU                       &                      & Dice                      & IoU                       \\ \hline
U-Net(MICCAI'15)                                                       & 0.512                     & 0.444                     &  & 0.710                     & 0.627                     &                      & 0.398                     & 0.335                     \\
U-Net++(TMI'19)                                                        & 0.483                     & 0.410                     &  & 0.707                     & 0.627                     &                      & 0.393                     & 0.335                     \\
PraNet(MICCAI'20)                                                      & 0.712                     & 0.640                     &  & 0.871                     & 9.797                     &                      & 0.628                     & 0.567                     \\
SA-Net(MICCAI'21)                                                      & 0.753                     & 0.670                     &  & 0.888                     & 0.815                     &                      & 0.750                     & 0.654                     \\
MSNet(MICCAI'21)                                                       & 0.751 & 0.671 &  & 0.865 & 0.799 &   & 0.723 & 0.652 \\
Polyp-Pvt(arXiv)                                                       & 0.808                     & 0.727                     &  & 0.900                     & 0.833                     &                      & 0.787                     & 0.706                     \\
\begin{tabular}[c]{@{}l@{}}BoxPolyp-Res2Net\\ (MICCAI'22)\end{tabular} & 0.820     & 0.741               &  & 0.903                     & 0.835                     &                      & 0.829                     & 0.742                     \\
LDNet(MICCAI'22)                                                       & 0.784                     & 0.706                     &  & -                     &-                    &                      & 0.744                     & 0.665                     \\
CL-Polyp(Ours)                                                         & 0.770                     & 0.710                     &  & 0.880                     & 0.812                     &                      & 0.760                     & 0.658                     \\ \hline
\end{tabular}}
\end{table}

\subsubsection{Results of Experiment Scenario III}
In this section, we show the necessity of each module in the network in the form of an ablation experiment.
We also showed the impact of different data enhancement methods on our approach.  We only use the Kvasir-SEG dataset to conduct experiments in this experimental part.

\begin{table}[]
\centering
\caption{{Performance} comparison of CL-Polyp with different modules on {the} Kvasir-SEG {dataset: Ablation study results}.}
\resizebox{0.7\linewidth}{!}{
\begin{tabular}{lcc}
\hline
Method                    & \multicolumn{2}{l}{Kvasir-SEG} \\ \cline{2-3} 
                          & Dice           & IoU           \\ \hline
Baseline                  & 0.889          & 0.827         \\
Baseline + MASPP          & 0.899          & 0.838         \\
Baseline + MASPP + CA     & 0.900          & 0.842         \\
Baseline + MASPP + CA + CL Branch & \textbf{0.918}    & \textbf{0.864}  \\ \hline
\end{tabular}}
\end{table}
\begin{table}[]
\centering
\caption{{Impact} of different data augmentation {techniques} on {CL-Polyp performance using the} Kvasir-SEG {dataset: Comparative results}.}
\resizebox{0.7\linewidth}{!}{
\begin{tabular}{lcc}
\hline
Method                                  & \multicolumn{2}{l}{Kvasir-SEG} \\ \cline{2-3} 
                                        & Dice          & IoU            \\ \hline
Base                                    & 0.911         & 0.856         \\
Base + GaussianBlur                     & \textbf{0.918}& \textbf{0.864} \\
Base + ToGray                           & 0.903         & 0.848          \\
Base + BrightnessContrast              & 0.906         & 0.851          \\
Base + GaussianBlur + BrightnessContrast & 0.910         & 0.855          \\
Base + GaussianBlur + RandomCrop(0.6 -1.0)& 0.909         & 0.855          \\
Base + GaussianBlur + GridDropout       & 0.909         & 0.855          \\ \hline
\end{tabular}}
\end{table}
We use the DeepLabV3+ network as our Baseline and verify effectiveness by adding components to Baseline. 
Baseline + MASPP + CA + CL branch is our final model, Cl-Polyp.

\textbf{Effectiveness of MASPP Module and CA Module.}
The results of Baseline + MASPP in Table 4 can demonstrate the effectiveness of the results. After adding the MASPP module, our approach showed a 1.0\% and 1.1\% improvement in Dice and IoU, respectively. After adding the CA module, our method showed a 0.1\% and 0.4\% improvement in Dice and IoU, respectively.

\textbf{Effectiveness of Contrastive learning branch.}
As shown in Table 4, after adding the CL branch, our approach has a 1.8\% and 2.2\% increase in Dice and IoU, respectively. This can fully shows that CL Branch has a nice effect.

\textbf{Effectiveness of different augmentations.}

Data enhancement methods are essential for learning data invariance by contrast learning, so we explored the influence of different data enhancement methods on our methods. 
Table 5 shows the experimental results, where Base data enhancement refers to random resize and crop, and the range of crop is (0.8, 1.0), random rotation, vertical flip, and normalization. 
Based on base data enhancement, we added gaussian blur,  to gray, brightness contrast, and their random combination and other contents. 
The ablation results show that {adding} only {Gaussian} blur to the Base data {augmentation} {yields} the best results.

\bibliographystyle{unsrt}  
\bibliography{templateArxiv}

\end{document}